\documentclass[numbib]{cta-author}

\usepackage{amssymb}
\usepackage{times}
\usepackage{epsfig}
\usepackage{graphicx}
\usepackage{epstopdf}
\usepackage{amsmath}
\usepackage{multirow}
\usepackage{enumerate}
\usepackage{booktabs}
\usepackage{bbm}
\usepackage[ruled,norelsize]{algorithm2e}
\SetKwComment{Comment}{$\triangleright$\ }{}
\makeatletter
\newcommand{\removelatexerror}{\let\@latex@error\@gobble}
\makeatother
\newcommand{\ra}[1]{\renewcommand{\arraystretch}{#1}}

{}
{}
{}

\begin{document}

\title{Multi-view Registration Based on Weighted Low Rank and Sparse Matrix Decomposition of Motions}

\author{\au{Congcong Jin$^{1,2}$, }
\au{Jihua Zhu$^{1\corr}$, }
\au{Yaochen Li$^1$, }
\au{Shanmin Pang$^1$, }
\au{Lei Chen$^3$, }
\au{Jun Wang$^4$}
}

\address{\add{1}{School of Software, Xi'an Jiaotong University, Xi'an, 710049}
\add{2}{State Key Laboratory of Rail Transit Engineering Informatization (FSDI), Xi'an, 710043}
\add{3}{School of Computer Science, Nanjing University of Posts and Telecommunications, Nanjing, 210003}
\add{4}{School of Digital Media, Jiangnan University, Wuxi, 214122}
\email{zhujh@xjtu.edu.cn}}

\begin{abstract}
Recently, the low rank and sparse (LRS) matrix decomposition has been introduced as an effective mean to solve the multi-view registration. It views each available relative motion as a block element to reconstruct one matrix so as to approximate the low rank matrix, where global motions can be recovered for multi-view registration. However, this approach is sensitive to the sparsity of the reconstructed matrix and it treats all block elements equally in spite of their varied reliability. Therefore, this paper proposes an effective approach for multi-view registration by the weighted LRS decomposition. Based on the anti-symmetry property of relative motions, it firstly proposes a completion strategy to reduce the sparsity of the reconstructed matrix. The reduced sparsity of reconstructed matrix can improve the robustness of LRS decomposition. Then, it proposes the weighted LRS decomposition, where each block element is assigned with one estimated weight to denote its reliability. By introducing the weight, more accurate registration results can be recovered from the estimated low rank matrix with good efficiency. Experimental results tested on public data sets illustrate the superiority of the proposed approach over the state-of-the-art approaches on robustness, accuracy, and efficiency.
\end{abstract}

\maketitle

\section{Introduction}
\label{intro}
Range scan registration has attracted broad interests due to its wide applications in robot mapping~\cite{Henry2012RGB,Borrmann2008Globally,N20076D}, 3D model reconstruction~\cite{Zhou2013Dense,Izadi2011KinectFusion}, object recognition~\cite{Abate07,Held16} and etc. The task of registration is to calculate the optimal transformation for two or more range scans so as to transfer them into one coordinate system and recover the original scene of a 3D object. Based on the number of scans to be registered, this problem can be classified into two categories: pair-wise registration and multi-view registration. The most popular method for pair-wise registration is the iterative closest point (ICP) algorithm proposed by Besl et al.~\cite{Besl92}, which iteratively builds up correspondences and calculates the optimal transformation by minimizing the residual error. Although the ICP algorithm has good performance in efficiency, it is a local convergent approach. Besides, this approach can not be applied to the registration of scan pair, which contains large non-overlapping areas. Therefore, a lot of ICP variants were proposed for the pair-wise registration.

To address non-overlapping areas, Chetverikov et al.~\cite{Chetverikova05} proposed the trimmed ICP (TrICP) algorithm, which introduced an overlap percentage into the original ICP algorithm. During each iteration, it requires to search an optimal overlap percentage, so it is time-consuming. Subsequently, Phillips et al.~\cite{Phillips07} proposed an efficient ICP variant called the fractional TrICP (FTrICP) algorithm, which can simultaneously compute the overlap percentage and rigid transformation for partially overlapping scans. For the local convergence issue, Fitzgibbon et al.~\cite{Fitzgibbon03} employed the Levenberg-Marquardt algorithm to expand the narrow convergent range of ICP algorithm. Besides, invariant features were introduced into ICP algorithm by Lee et al.~\cite{Sharp2002ICP}. Moreover, the genetic algorithm~\cite{Lomonosov2006Pre,Zhu2014Robust} and particle filter~\cite{Sandhu2010Point} were utilized to search the optimal rigid transformation. To boost the accuracy, some probabilistic methods ~\cite{Granger02,Jian11,Myronenko10,Tsin04} were also proposed. Much precise these methods may be, the huge computational resources they require poses a great challenge to most application areas. To improve the robustness, some methods have also been investigated in recent years. Based on the ratio of bidirectional distances, Zhu et al.~\cite{Zhu2016Registration} proposed an ICP variant, which assigns a probability for each correspondence. Besides, Xu et al. ~\cite{Hasanbelliu2014Information} introduced the concept of correntropy into pair-wise registration and proposed the approach to achieve the pair registration by maximizing the correntropy of one scan pair.

Although these approaches may obtain good results for pair-wise registration, they are not suitable for the multi-view registration. Therefore,
many researchers explore the principle of pair-wise registration and extend it to solve the multi-view registration. The original approach was proposed
by Chen et al.~\cite{chen1992object}, which repeatedly aligns two scans and merge them into one model until all range scans are integrated into the whole model. This approach is simple and efficient, but it suffers from the problem of error accumulation. To address this issue, Bergevin et al.~\cite{bergevin1996towards} proposed an ICP based registration approach, which can simultaneously align one scan to all the other scans. Since this approach should establish point correspondences between one scan and all the others, it is very time-consuming. Subsequently, the multi-z-buffer technique~\cite{benjemaa1999fast} was then introduced into multi-view registration to improve the efficiency. To further reduce the accumulative error, some other approaches~\cite{shih2008efficient,torsello2011multiview} view multi-view registration as the optimization problem over the graph of adjacent scans, which transfer the registration error between coordinate systems. As these approaches do not need to update the point correspondences during registration, they cannot really reduce accumulative error, but just distribute it over all scans.

Recently, Mateo et al. \cite{mateo2014bayesian} utilized the Bayesian framework to deal with missing data of pair-wise correspondences in multi-view registration, which then be solved by the Expectation-Maximization algorithm. Related works can also be found in~\cite{evangelidis2014generative}. Besides, Toldo et al. ~\cite{toldo2010global} proposed an ICP and Generalized Procrustes Analysis~\cite{beinat2001generalised} combined approach to achieve multi-view registration. To explore the redundant information of non-adjacent range scans, Godvin et al.~\cite{govindu2014averaging} introduced the Lie-Algebraic averaging~\cite{govindu2004lie} algorithm to refine global motions. This approach was then extended by Li et al.~\cite{Zhongyu2014Improved} to achieve more accurate and efficient multi-view registration. Besides, Guo et al.~\cite{Guo2017Weighted} proposed a weighted motion averaging algorithm to improve the  accuracy of multi-view registration. More recently, Arrigoni et al. \cite{arrigoni2016global} cast the multi-view problem into the framework of the low-rank and sparse (LRS) matrix decomposition. By decomposing the relative motion stacked matrix, the noise matrix is discarded and low rank matrix can be obtained to recover global motions for multi-view registration. However, this approach is sensitive to the sparsity of the stacked matrix to be decomposed. Besides, it treats all relative motions equally in spite of their varied reliabilities, which is not good for multi-view registration.

In this paper, we extend the approach presented in~\cite{arrigoni2016global} to achieve more effective registration of multi-view range scans. The contribution of this paper can be delivered as follows: a matrix completion strategy is proposed to reduce the sparsity of potentially decomposed matrix based on the anti-symmetry property of relative motions (block elements). Then, a weight value is estimated and assigned to denote the reliability of each non-zero block elements. Moreover, the L1-ALM~\cite{zheng2012practical} algorithm is extended to decompose the weighted matrix and obtain accurate global motions for multi-view registration. To demonstrate its effectiveness, the proposed approach was also tested on public available data sets.

The rest of this paper is organized as follows: Section~\ref{LRSmethod} briefly introduces the LRS decomposition framework for the multi-view registration.
Then in Section~\ref{proposed_algo}, the proposed approach is presented in details. Following that is Section~\ref{experiments}, in which the proposed approach was tested and compared with some related approaches. Finally, some conclusions and future works are drawn in Section~\ref{conclusions}.

\section{LRS decomposition for multi-view registration}\label{LRSmethod}

Suppose there are $N$ range scans, which are acquired from one object in different views.
Let ${M_i} \in SE(3)$ be the global motion, which denotes the rigid transformation between the local reference
frame of the $i$th range scan and the global coordinate system:
\begin{align}
{M_i} = \left[ {\begin{array}{*{20}{c}}
   {{{\mathop{\rm R}\nolimits} _i}} & {{{\vec t}_i}}  \\
   0 & 1  \\
\end{array}} \right],
\label{eq: 7}
\end{align}
where ${\bf{R}_i} \in SO(3)$ and ${\vec t_i} \in \mathbb{R}^3$ represent the rotation matrix and the translation vector, respectively.
Obviously, the rank of $M_i$ is 4. Given initial global motions, the task of multi-view registration is to estimate accurate global motions
for all range scans. Without loss of generality, the global coordinate system can be attached to the local reference frame of the first range scan. Related to global motion, there is another kind of motion called as the relative motion ${M_{ij}}$,  which represents the rigid transformation between the reference frame of the $i$th scan and that of the $j$th scan:
\begin{align}
{M_{ij}} = M_i^{ - 1}{M_j}
\label{eq: Rmo}
\end{align}
where $M_i^{ - 1}$ denotes the inverse of motion $M_i$ and the rank of $M_{ij}$ is also 4.

For the multi-view registration, Arrigoni et al. ~\cite{arrigoni2016global} proposed the LRS matrix decomposition based approach. Before presenting this approach, three block matrix should be introduced and defined as:
\begin{align}
V = [{M_1}\quad {M_2}\quad  \cdots \quad {M_N}],\quad {U} = \left[ {\begin{array}{*{20}{c}}
   {M_1^{ - 1}}  \\
   {M_2^{ - 1}}  \\
    \vdots   \\
   {M_N^{ - 1}}  \\
\end{array}} \right]{\rm{ }}
\end{align}
and
\begin{align}
\begin{split}
X &=\begin{bmatrix}
    {I_4}      & M_{12}  & \dots  & M_{1N} \\
    M_{21} & {I_4}       & \dots  & M_{2N} \\
    \vdots & \vdots           & \ddots & \vdots \\
    M_{N1} & M_{N2}           & \dots  & {I_4}
\end{bmatrix}\\
\end{split}
\label{eq: 10}
\end{align}
where $I_4$ indicates the $4 \times 4$ identity matrix.
Therefore, $X=UV$ and $U^TU= I_4$. Although the matrix $X$ is larger than $M_i$ or $M_ij$, they have the same rank due to
the special structure of $X$.

For the multi-view registration, the LRS decomposition based approach views each available relative motions as a block element to reconstruct the matrix $\hat X$,
where the non-available relative motions are replaced by zero matrix. As the reconstructed matrix $\hat X$ is the approximation of $X$, they have the following relation:
\begin{align}
\hat X = X + E
\end{align}
where $E$ is called as error matrix, which is a sparse matrix containing noises and outliers.

According to \cite{arrigoni2016global}, the multi-view registration appcan be formulated as the following optimization problem:
\begin{align}
\begin{array}{l}
 \mathop {\min }\limits_{X} \left\| {{\mathcal {P}_\Omega }(X - \hat X)} \right\|_F^2 \\
 {\rm{s}}{\rm{.t}}{{. \quad  X = UV, \hat X =X + E}} \\
 \end{array}
 \label{eq: LRS}
 \end{align}
where ${\mathcal {P}_\Omega }(X - \hat X)$ represents the projection of $(X - \hat X)$ onto $\Omega$ and $\Omega$ is an indicator matrix indicating whether the corresponding block element in the reconstructed matrix $\hat X$ is available or not. Subsequently, the LRS decomposition algorithm can be applied to solve Eq. (\ref{eq: LRS}) and obtain the matrix $X$, which is used to recover global motions for muti-view registration.

Although the framework of LRS decomposition for multi-view registration has been proposed in \cite{arrigoni2016global}, this approach is sensitive to the sparsity of reconstructed matrix. Besides, it treats all block elements equally in spite of their varied reliability. To obtain good registration results, more effective approach is required.

\section{The proposed approach}\label{proposed_algo}

Although the LRS decomposition has been introduced to solve the multi-view registration \cite{arrigoni2016global}, two limitations should be stated. Firstly, this approach is sensitive to the sparsity of reconstructed matrix to be decomposed.
Then, during matrix decomposition, it treats all blocking elements equally and does not take varied reliability of each relative motion into consideration, which may lead to unexpected registration results.
To address these two issues, we propose an effective LRS decomposition based approach for multi-view registration and its flowchart is displayed in Fig. 1.
As shown in Fig. 1, the proposed approach consists of the following four major steps, which will be presented with more details.

\subsection{Matrix reconstruction}

For the matrix reconstruction, it is required to obtain relative motions, which can be estimated by the pair-wise registration approach.
By consideration both the efficiency and accuracy, the trimmed ICP algorithm is utilized to estimate the relative motion of scan pair.

Suppose there are two partially overlapping range scans in $\mathbb{R}{^3}$, a
data shape $P \buildrel \Delta \over = \{ {\vec p_a}\} _{a =
1}^{{N_p}}$ and a model shape $Q \buildrel \Delta \over = \{ {\vec
q_b}\} _{b = 1}^{{N_q}}$ $({N_p},{N_p} \in \mathbb{N})$. Given initial rigid transformation, the TrICP algorithm achieves the pair-wise registration by minimizing the following objective function:
\begin{equation}
\begin{array}{l}
 \psi (\xi ,M) = \frac{1}{{\left| {{P_\xi }} \right|{\xi ^{1 + \lambda }}}}\sum\limits_{{{\vec p}_a} \in {P_\xi }} {\left\| {{\bf{R}}{{\vec p}_a} + \vec t - {{\vec q}_{c(a)}}} \right\|_2^2}  \\
 {\rm{s}}{\rm{.t}}{\rm{.}}\quad {{\bf{R}}^T}{\bf{R}}{\rm{ = }}{{\rm{I}}_3},\quad \det ({\bf{R}}) = 1 \\
 \end{array}
\label{eq: TrICPObj}
\end{equation}
where $\xi $ denotes the overlap percentage parameter, ${{P_\xi }}$ represents the overlapping part of data shape $P$ to model shape $Q$,  $\left|  \cdot  \right|$ represents the cardinality of set, ${{{\vec q}_{c(a)}}}$ is the correspondence of ${{{\vec p}_a}}$ and $\lambda$ ($\lambda$=2) is a preset parameter.

\begin{figure*}[htb!]
\begin{center}
\centerline{\includegraphics[scale=0.6]{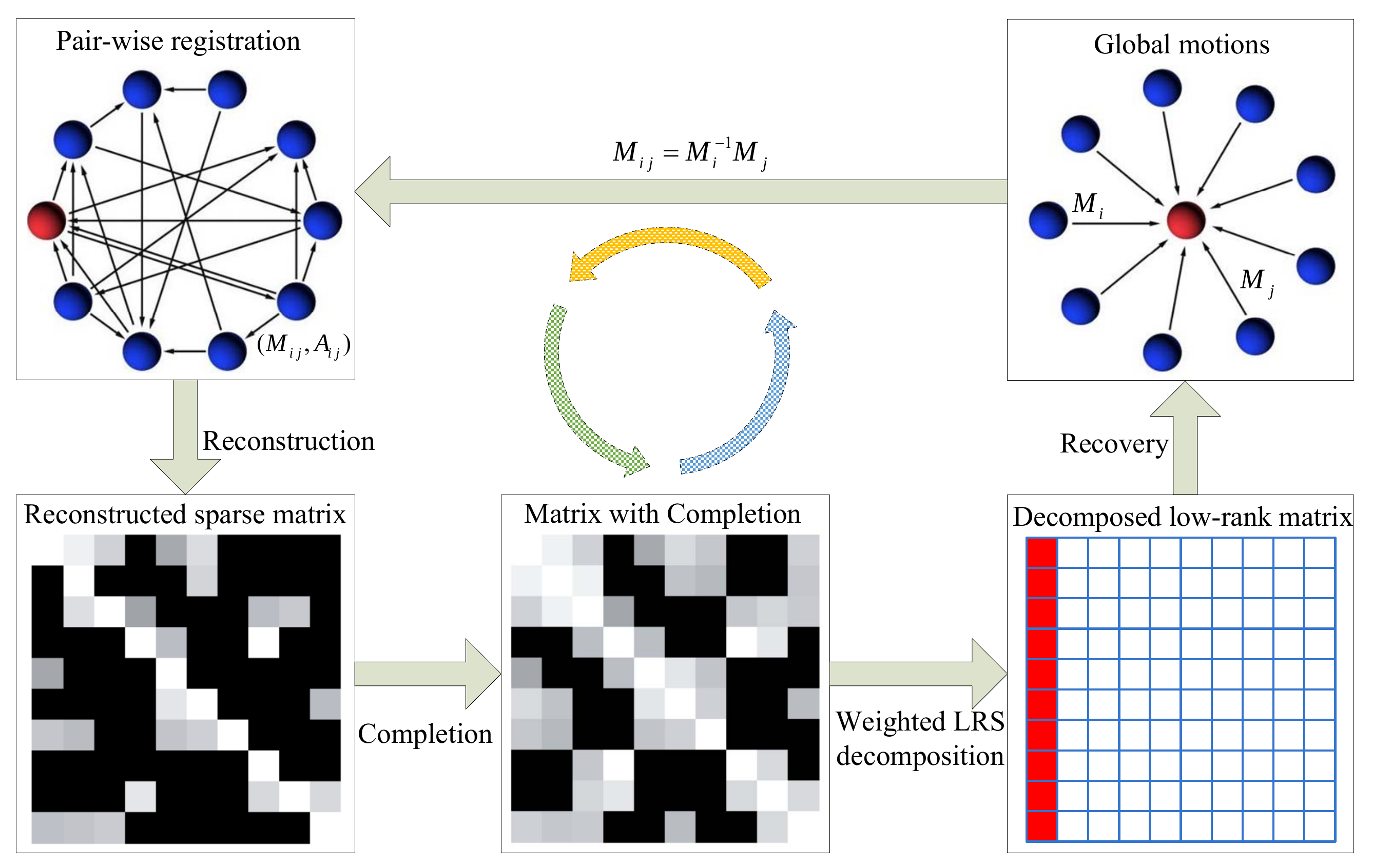}}
\end{center}
   \caption{The flowchart of the proposed approach. In the 1st row, each dot denotes one range scan, the red dot is the rang scan attached with the global coordinate system and each line with arrow indicates one motion. In the 2nd row, each square denotes one relative motion and its reliability is indicated by the gray value, where the black one is unobserved and the white one is very reliable. Besides, the column of red squares denote block elements used for the recovery of global motions.}
\label{fig:LRSFrame}
\end{figure*}

\begin{figure*}[htbp]
\centering
\begin{minipage}[b]{0.33\linewidth}
  \centering
  \centerline{\includegraphics[scale=0.6]{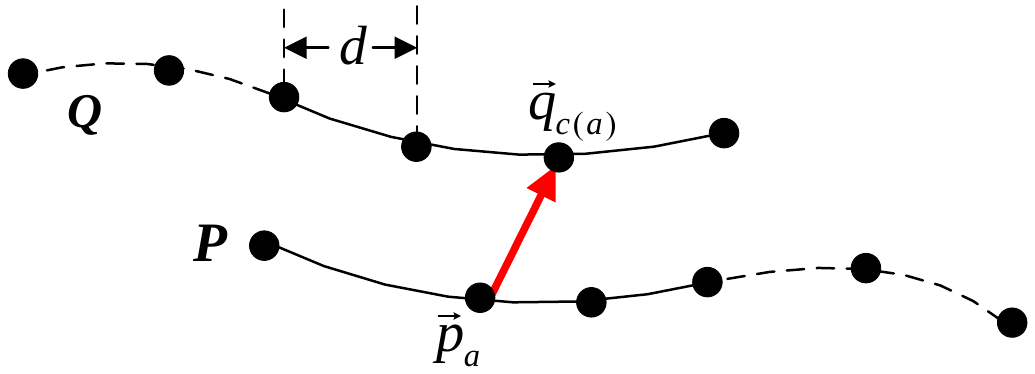}}
  \centerline{(a)}
\end{minipage}
\begin{minipage}[b]{0.33\linewidth}
  \centering
  \centerline{\includegraphics[scale=0.6]{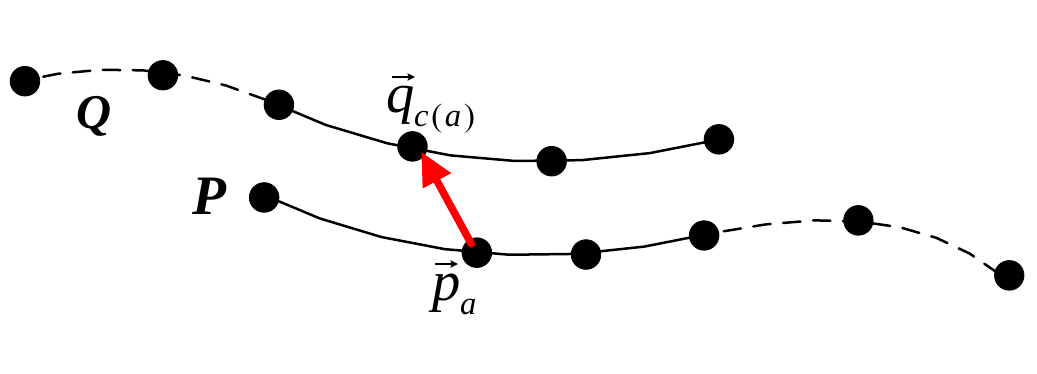}}
  \centerline{(b)}
\end{minipage}
\begin{minipage}[b]{0.33\linewidth}
  \centering
  \centerline{\includegraphics[scale=0.6]{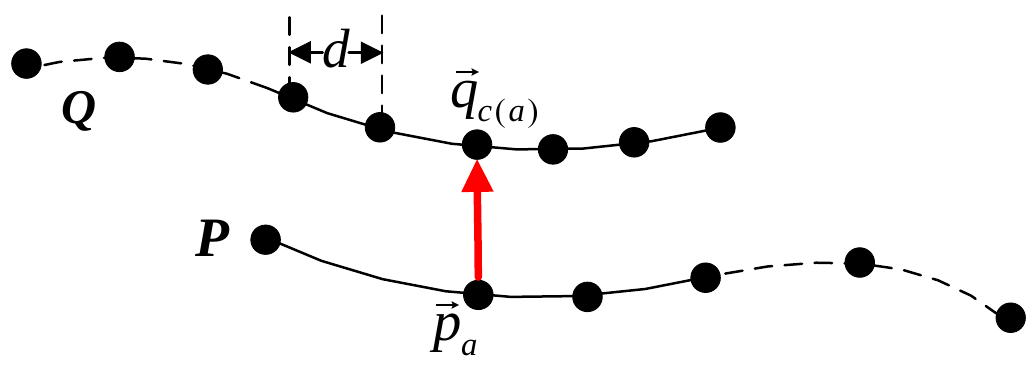}}
  \centerline{(c)}
\end{minipage}
\caption{ Illustration of some factors related to the trimmed MSE. (a) Large trimmed MSE caused by inaccurate registration and low resolution of model shape. (b) Accurate registration leads to small trimmed MSE. (c) High resolution of model shape can reduce the trimmed MSE.}\medskip
\label{fig:w}
\end{figure*}

\subsubsection{Estimation of overlapping percentage}

Although the TrICP algorithm is effective for pair-wise registration, it only suitable for the registration of scan pair with a certain amount of overlap percentages. To obtain reliable relative motions, the TrICP algorithm can only applied to these scan pairs that satisfy ${\xi _{ij}} > {\xi _{thr}}$, where ${\xi _{thr}}$ is a predefined threshold. Therefore,
it is required to estimate the overlap percentage for each scan pair before pair-wise registration. To address this issue, the method proposed in~\cite{Zhongyu2014Improved} is directly utilized. For each point in the $i$th range scan, it firstly searches correspondences from each other scans. According to their distances, these point pairs can be sorted in ascending order.
By traversing each sorted point pair, all its front point pairs can be used to calculate the value of objective function (\ref{eq: TrICPObj}). The distance of the point pair, which minimizes the objective function, can be viewed as the distance threshold.
For the $j$th range scan, if there are $n_{ij}$ point pairs, whose distances are smaller than the distance threshold, then the overlap percentage $\xi_{ij}$ is estimated as follows:
\begin{align}
\xi_{ij} = \frac{{n_{ij}}}{{n_i}}
\label{eq:overpercen}
\end{align}
where $n_i$ denotes the number of point in the $i$th range scan. It should be noted that
$\xi_{ij}$ and $\xi_{ji}$ are two different overlap percentages, which are always unequal.
For each scan pair, if its overlap percentage satisfies ${\xi _{ij} > {\xi_{thr}}} $, then the $i$th range scan and the $j$th range scan can be viewed as the data shape and model shape, respectively. Further, the TrICP algorithm can be directly used to estimate its relative motion $M_{ij}$. Here, $\xi _{thr}= 0.4$ can guarantee the TrICP algorithm to achieve reliable pair-wise registration.

\subsubsection{Estimation of weight}

Actually, the reliability of each relative motion are varied due to many reasons, such as noise level, resolution of range point and overlap percentage of scan pair.
To indicate its reliability, a weight requires to be estimated for each relative motion. Intuitively, the smaller the trimmed mean square error (MSE) is, the more reliable the relative motion is. Before introducing the estimation method, some factors related to the trimmed MSE should be presented and analyzed. As shown in Fig. \ref{fig:w}, accurate pair-wise registration will lead to small trimmed MSE. However, the trimmed MSE also related to the point resolution of model shape. More specifically, with the same registration accuracy, high resolution of model shape will lead to small trimmed MSE. Vice vera. Accordingly, a weight $A_{ij}$ for the relative motion $M_{ij}$ can be reasonably estimated as follows:
\begin{align}
{A_{ij}} = \frac{{Qe}}{{Pe^2}}
\label{eq: defA}
\end{align}
where $Me$ indicates the point resolution of model shape and $Pe$ denotes the trimmed MSE of aligned scan pair, which can be directly obtained by the TrICP algorithm. More specifically,
they can be calculated as follows:
\begin{align}
\begin{split}
&Qe = \frac{1}{N_q}(\sum\limits_{i = 1}^{{N_q}} {d_{_{{{\vec q }_i}{{\vec q }_{c(i)}}}}^2} )\\
&Pe = \frac{1}{|{P_\xi }|}(\sum\limits_{i = 1}^{\left| {{P_\xi }} \right|} {d_{{{\vec p }_i}{{\vec q }_{c(i)}}}^2} )
\end{split}
\label{eq: 14}
\end{align}
where $d_{{{\vec p }_i}{{\vec q }_{c(i)}}}$ denotes the distance of one point pair located in the overlapping areas, $d_{_{{{\vec q }_i}{{\vec q }_{c(i)}}}}$ represents the distance of one point ${\vec q _i}$ in the model shape to its nearest neighbor ${\vec q _{c(i)}}$ in the model shape itself. According to the definition of weight, reliable relative motion will lead to large weight.

As there maybe large difference between varied relative motions, it is better to normalize all weights as follows:
\begin{align}
{B_{ij}} = \frac{{{A_{ij}}}}{ \max (A)}
\label{eq: Norm}
\end{align}
where ${ \max (A)}$ denotes the maximum value in $A$. It should also be noticed that this is the weight for each relative motion.
To associate with each element of ${M_{ij}}$, the following expanding is required:
\begin{align}
\begin{split}
{({W_{ij}})_{4 \times 4}} = {B_{ij}} \otimes {\mathbbm{1}_{4 \times 4}}
\end{split}
\label{eq: 15}
\end{align}
where $ \otimes $ indicates the Kronecker product and ${\mathbbm{1}_{4 \times 4}}$ denotes a $4 \times 4$ matrix filled by ones.

After the pair-wise registration, the relative motion $M_{ij}$ can be utilized to
reconstruct the $\hat X$, which is viewed as the approximation of the low rank matrix $X$.
In the same way, its corresponding weight $W_{ij}$  can be utilized to
reconstruct the matrix $W$.

\subsection{Matrix Completion}
As the LRS decomposition algorithm is sensitive to the sparsity of reconstructed matrix to be decomposed, it is better to obtain as many relative motions as possible so as to reconstruct a matrix with reduced sparsity. Therefore, it is required to design a completion strategy for the matrix reconstruction.

According to Eq. (\ref{eq: Rmo}), the relative motion $M_{ji}$ can be calculated as:
\begin{align}
{M_{ji}} = M_j^{ - 1}{M_i}
\label{eq: Rmo1}
\end{align}
Obviously, there exits the property of anti-symmetry between the pair of relative motions $(M_{ij}, M_{ji})$, which is formulated as:
\begin{align}
{M_{ji}} = M_{ij}^{-1}
\label{eq: Rmo2}
\end{align}
where $M_{ij}^{ - 1}$ indicates the inverse of motion $M_{ij}$. For a pair of scans acquired from one object, two different overlap percentages $(\xi_{ij}, \xi_{ji})$ are estimated by the proposed method.
Suppose one of them is larger than the predefined threshold $\xi_{thr}$ and the other is smaller than $\xi_{thr}$. Subsequently, one relative motions
with a weight can be available to reconstruct the $\hat X$. To reduce the sparsity of $X$, the anti-symmetry property of relative motion can be applied to
obtain the other relative motion with its weight assigned as:
\begin{align}
{W_{ji}} = W_{ij}
\label{eq: w}
\end{align}

As shown in Fig. 1, this matrix completion strategy can seriously reduce the sparsity of reconstructed matrix to be decomposed and will lead to robust results of LRS matrix decomposition.

\subsection{Weighted LRS decomposition}

After matrix completion, the matrix $\hat X$ has been reconstructed with less missing data. Moreover, the weight matrix is also provided to indicate the reliability of component at the same position in $\hat X$. Therefore, the weighted LRS decomposition should be designed to approximate the low rank matrix $X$.

According to \cite{zheng2012practical}, it is reasonable to use the robust $L$1-norm as the measurement to approximate $X$. The approximation can be formulated as the optimization problem:
\begin{equation}
\begin{aligned}
& \underset{U,V,E}{\text{min}}
& & {||W\odot E||_1+\lambda ||X||_*} \\
& \text{s.t.}
& &  {\hat X}{=X + E}
\end{aligned}
\label{eq: wLRS1}
\end{equation}
where $W$ is a weight matrix, which indicates the reliability of component at the same position in $\hat X$.
As mentioned before, $X=UV$ and $U^TU= I_4$. Therefore, $||X||_*=||V||_*$. By replacing the trace-norm regularizer,
Eq. (\ref{eq: wLRS1}) can be reformulated as:
\begin{equation}
\begin{aligned}
& \underset{U,V,E}{\text{min}}
& & {||W\odot E||_1+\lambda ||V||_*} \\
& \text{s.t.}
& &  {\hat X}{=UV + E}, {{U^TU}{=I_4}}
\end{aligned}
\label{eq: wLRS2}
\end{equation}
After the investigation of this problem, we find that the Augmented Lagrange Multiplier (ALM) method can be utilized to solve it.

Benefited from the ALM algorithm, the corresponding augmented Lagrange function can be derived as:
\begin{align}
\begin{split}
f(U,&V,E,L,\mu ) = {\left\| {W \odot E} \right\|_1} + \lambda {\left\| V \right\|_*} + \\
&\left\langle {L,\hat X - UV - E} \right\rangle  + \frac{\mu }{2}\left\| {\hat X - UV - E} \right\|_F^2
\end{split}
\label{eq: 17}
\end{align}
where $L$ is the Lagrange multiplier, $\mu $ denotes the penalty parameter and $\left\langle {.,.} \right\rangle $ indicates the inner product of two matrices.
As it is somewhat same to the problem discussed in ~\cite{zheng2012practical}, this problem can also be solved the Gauss-Seidel Iteration algorithm. In each iteration, three steps are alternately utilized to estimate $U$, $V$ and $E$, respectively.

\subsubsection{Solving $U$ via Orthogonal Procrustes}

By fixing $E$ and $V$, the update of $U$ can be simplified as the following problem:
\begin{equation}
\begin{aligned}
& \underset{U}{\text{min}}
& & {\frac{\mu }{2}\left\| {(\hat X - E + \frac{1}{\mu }L}) - UV  \right\|_F^2} \\
& \text{s.t.}
& & {\hat X}{=UV + E}, {{U^TU}{=I_r}}
\end{aligned}
\label{eq: 18}
\end{equation}

The above orthogonal procrustes problem can be solved by the SVD method of $(\hat X - E + \frac{1}{\mu }L){V^T}$:
\begin{align}
\begin{split}
\left[ {{U_1}\quad{\rm{ }}{S_1}\quad{\rm{ }}{V_1}} \right] = svd((\hat X - E + \frac{1}{\mu }L){V^T})
\end{split}
\label{eq: 19}
\end{align}
Consequently, $U$ can be derived as:
\begin{align}
\begin{split}
U = {U_1}V_1^T
\end{split}
\label{eq: 20}
\end{align}

\subsubsection{Solving $V$ via Singular Value Decomposition}

Given $E$ and $U$, the update of $V$ can be achieved by solving:
\begin{align}
\begin{split}
\mathop {\min }\limits_V  \{{\rm{ }}\lambda {\left\| V \right\|_*} +\left\langle {L,\hat X - E - UV} \right\rangle  + \frac{\mu }{2}\left\| {\hat X - E - UV} \right\|_F^2\}
\end{split}
\label{eq: 21}
\end{align}
Since ${U^T}U = {I_r}$, the above objective function can be reformulated as:
\begin{align}
\begin{split}
\mathop {\min }\limits_V \{{\rm{ }}\lambda {\left\| V \right\|_*} +\frac{\mu }{2}\left\| {V -U^T ( \hat X- E + \frac{1}{\mu }L)} \right\|_F^2\}
\end{split}
\label{eq: 22}
\end{align}

\begin{figure}[t]
\removelatexerror
\begin{algorithm}[H]
\SetKwInOut{KwIn}{Input}
\SetKwInOut{KwOut}{Output}
\KwIn{$X$, $W$, $L$, $\mu$, and $\rho = 1.05$}
\KwOut{$U$ and $V$}
\caption{ Weighted LRS decomposition algorithm}
\Comment*[h]{{\fontfamily{ptm}\selectfont \emph{Outer Iteration}}}\\
\While{not converged}
{
   \Comment*[h]{{\fontfamily{ptm}\selectfont \emph{Inner Iteration}}}\\
   \While{not converged}
   {
      Update $U$ according to Eq. (\ref{eq: 19}) and (\ref{eq: 20});\\
      Update $V$ according to Eq. (\ref{eq: 24}) and (\ref{eq: 25});\\
      Update $E$ according to Eq. (\ref{eq: E1}) and (\ref{eq: E2});
   }
   Update $L$ via $L = L+ {\mu}(M-E-UV)$;\\
   Update $\mu$ via ${\mu} =$ min $\left \{{{{\rho}\mu}, 1e20} \right \}$.
}
\label{algo:ALM}
\end{algorithm}
\end{figure}

To solve problem (\ref{eq: 22}), the soft-thresholding (shrinkage) operator is adopted:
\begin{align}
\begin{split}
{\mathbf{S}_\varepsilon }\left[ q \right] = max(\left| q \right| - \varepsilon ,0)sgn(q)
\end{split}
\label{eq: 23}
\end{align}
where $sgn(q)$ is the sign function. Then the singular values of ${U^T}(\hat X - E + \frac{1}{\mu }L)$ is computed as:
\begin{align}
\begin{split}
[{U_2}\quad{\rm{  }}{S_2}\quad{\rm{  }}{V_2}] = svd({U^T}(\hat X -E + \frac{1}{\mu }L))
\end{split}
\label{eq: 24}
\end{align}

Finally, the shrinkage operator is applied to the singular values and the optimal $V$ can be updated as:
\begin{align}
\begin{split}
V = {U_2}{\mathbf{S}_{\frac{\lambda }{\mu }}}[{S_2}]V_2^T
\end{split}
\label{eq: 25}
\end{align}

\subsubsection{Solving $E$ via Absolute Value Shrinkage}

Provided with $U$ and $V$, the error matrix $E$ can be updated as follows:
\begin{align}
\mathop {\min }\limits_E {\rm{ }}{\left\| {W \odot E} \right\|_1} + \frac{\mu }{2}\left\| {\hat X - E - (UV - \frac{1}{\mu }L)} \right\|_F^2
\label{eq: 26}
\end{align}

Since some block element of $\hat X$ is unobservable, the update of $E$ should be divided into two parts. Corresponding to the observable part of $\hat X$, the elements of $E$ can be updated by the absolute value shrinkage:
\begin{align}
\Omega \odot E =  \Omega \odot {\mathbf{S}_{\frac{W}{\mu }}}\left[ {\hat X - UV + \frac{1}{\mu }L} \right].
\label{eq: E1}
\end{align}
where $\Omega = \left\lceil W \right\rceil$ and $\left\lceil W \right\rceil$ denotes the ceiling operation of each element of $W$. While, elements corresponding to missing entries of $\hat X$ should be updated by:
\begin{align}
\bar \Omega \odot E= \Omega  \odot (X - UV + {\mu ^{ - 1}}L)
\label{eq: E2}
\end{align}
where $\bar \Omega$ denotes the complement of $\Omega$ and $\bar \Omega= ({\mathbbm{1}_{N \times N}}- \Omega)$.

The weighted LRS decomposition algorithm is summarized in Algorithm~\ref{algo:ALM}. By the application of this algorithm, two matrix $U$ and $V$ are obtained to approximate $X= UV$.

\subsection{Recovery of global motions}

After the LRS decomposition, we obtain $X = UV$. Theoretically, these block elements located in the first column of $X$ can directly be viewed as global motions for multi-view registration. However, these block elements may not be the elements of Special Euclidean groups $ SE(3)$ due to no constraint imposed on matrix decomposition. Accordingly, some operations are required to recover each global motion $M_i$ from one block elements located in the first column of $X$. Firstly, the corresponding block element is assign to $M_i=X(1:4,4i-3:4i)$, then $M_i$ is normalized by its element $M_i(4,4)$ as ${M_i} = {{{M_i}} \mathord{\left/{\vphantom {{{M_i}} {{M_i}(4,4)}}} \right. \kern-\nulldelimiterspace} {{M_i}(4,4)}}$. Besides, three elements $M_i(4,1:3)$ should be assigned with the zero value. Finally, the Singular Value Decomposition (SVD) can be applied to $M_i$ as follows:
\begin{align}
{{ }}{M_i}{{(1:3,1:3) = U_3}} Q {V_3^T}
\end{align}
where
\begin{align}
{{[U_3,S_3,V_3] = svd(}}{M_i}{{(1:3,1:3))}}
\end{align}
and $Q$ represents a $3 \times 3$ diagonal matrix with the elements of $(1, 1, det(U_3V_3^T) )$ on the main diagonal. After these operations, each block element located in the first column of $X$ can be recovered as one global motion for muti-veiw registration.

\subsection{Implementation details}

Accordingly, the overall process of the proposed multi-view registration approach is summarized in Algorithm 2.

\begin{figure}[!t]
\removelatexerror
\begin{algorithm}[H]
\SetKwInOut{KwIn}{Input}
\SetKwInOut{KwOut}{Output}
\KwIn{Multi-view range scans $\left\{ {{S_1},{S_2},...,{S_N}} \right\}$ and initial global motions $\left\{ {I,{{\hat M}_2},...,{{\hat M}_N}} \right\}$}
\KwOut{Accurate global motions $\left\{ {I,{M_2},...,{M_N}} \right\}$}
\caption{The proposed multi-view registration approach}
\Comment*[h]{{\fontfamily{ptm}\selectfont \emph{Outer iteration}}}\\
\While{not converged}
{
   Estimate overlap percentages for all scan pairs by Sec. 3.1.2;\\
   Select scan pairs that satisfy $\xi _{ij} > \xi _{thr}$;\\
   \Comment*[h]{{\fontfamily{ptm}\selectfont \emph{Inner iteration}}}\\
   \For(){Each selected scan pair}
   {
      Refine its relative motions by TrICP algorithm;\\
      Compute its weights by Sec. 3.1.3;\\
   }
   Reconstruct $\hat X$ and $W$ using available $M_{ij}$ and $W_{ij}$;\\
   Complete $\hat X$ and $W$ by Eqs. (14) and (15).;\\
   Approximate $X=UV$ according to Algorithm~\ref{algo:ALM};\\
   Recover global motions by Sec. 3.4;
}
\label{algoProcess}
\end{algorithm}
\end{figure}

With the matrix completion, the proposed approach can improve the robustness and efficiency of the LRS decomposition for multi-view registration. By proposing the weighted LRS decomposition, the accuracy of multi-view registration can be increased.

\section{Experiments}\label{experiments}

To show the performance of the proposed approach, experiments were conducted on seven data sets from the Stanford 3D Scanning Repository~\cite{Stanford} and UWA 3D Modeling Dataset~\cite{UWA}. Registration results are reported in the form of the objective function value ($Obj.$), which was designed in ~\cite{Zhu2014Surface}. As all multi-view registration approaches can estimate the optimal global motion
${{{M}}_{global}} = \{ {{I}},{{{M}}_2},{{{M}}_3}, \cdots ,{{{M}}_N}\}$, it is convenient to define the operation ${{M}_i} \oplus {P_i} = \{ {{{R}}_i}{\vec p_a} + {\vec t_a}\} _{a = 1}^{{N_i}}$ and reconstruct the integrated model:
\begin{align}
P = \{ {P_1},{{{M}}_2} \oplus {P_2},......,{{{M}}_N} \oplus {P_N}\}.
\end{align}
Then, one special model is defined for each scan ${P_i}$ as follows:
\begin{align}
{Q_i} = P\backslash ({{\bf{M}}_i} \oplus {P_i}) \buildrel \Delta \over = \{ {\vec q_b}\} _{b = 1}^{(\sum\nolimits_{j = 1}^N {{N_j}}  - {N_i})}.
\end{align}
For the accuracy evaluation, the $Obj.$ is calculated as follows:
\begin{align}
Obj. = \frac{1}{N}\sum\limits_{i = 1}^N {\psi ({\xi _i},{{{M}}_i})},
\end{align}
where $\psi ( \cdot )$ denotes the function displayed in Eq. (\ref{eq: TrICPObj}). To establish point correspondences, the $k-d$ tree based method was adopted to search the nearest-neighbor. All codes were implemented in Matlab on a desktop with four-core 3.6GHz processor and 8GB of memory.

\subsection{Validation}

To validate the proposed approach, it are compared with three versions of LRS decomposition based multi-view registration approaches: original LRS decomposition (LRS), LRS with matrix completion (LRS with MC), weighted LRS decomposition (Weighted LRS). For each data set, the same initial parameters were provided for four approaches. Registration results are shown in the form of $Obj.$ and run time. Tab \ref{tab:validation} displays registration results for all LRS decomposition based approaches. As shown in Tab \ref{tab:validation}, the original LRS decomposition may get the worst results for multi-view registration. Compared with the original LRS decomposition, the introduction of matrix completion and motion weight can both improve the performance of multi-view registration. Moreover, the integration of matrix completion and motion weight leads to development of the proposed approach, which can always obtain the best results for multi-view registration.

As stated before, the matrix for LRS decomposition is reconstructed by available relative motions, which are estimated by the pair-wise registration approach. However, most pair-wise registration approaches are unable to obtain reliable results for scan pairs with low overlap percentage, so the reconstructed matrix are always sparse. Since the LRS decomposition is sensitive to the sparsity of the reconstructed matrix, it is difficult to obtain good results. To reduce the sparsity, some block elements of the reconstructed matrix can be completed due to the anti-symmetry property of relative motions. With the reduced sparsity, the robustness of LRS decomposition is increased, which can lead to good multi-view registration. As relative motions of scan pairs are estimated by the pair-wise registration, their reliability are varied due to different overlap percentages of scan pairs. In the original LRS decomposition, the varied reliability is ignored, which is harmful for multi-view registration. By the analysis of pair-wise registration, we use the trimmed MSE and resolution of model shape to calculate the weight, which denotes the reliability of each relative motion. By introducing the weight in LRS decomposition, the accuracy of multi-view registration is increased. What's more, the combination of matrix completion and motion weight can further improve the performance of LRS decomposition for multi-view registration.

\begin{table*}[htbp]
\caption{Comparison of different LRS decomposition based approaches, where small value indicates good performance and bold number denotes the best result.}
\label{tab:validation}
\ra{1.3}
\scalebox{1}{
\begin{tabular}{@{}llrrrrrcrrrrrcrrrrrcrrrrr@{}}%
\toprule
&&& \multicolumn{5}{c}{LRS} && \multicolumn{5}{c}{LRS with MC} &&\multicolumn{5}{c}{Weighted LRS} &\multicolumn{5}{c}{Ours}\\
\cmidrule{5-7} \cmidrule{11-13} \cmidrule{17-19} \cmidrule{23-25}
Datasets &&&& \emph{Obj.} &&\emph{T(s)} &&&& \emph{Obj.} && \emph{T(s)} &&&&  \emph{Obj.} && \emph{T(s)} &&&& \emph{Obj.} && \emph{T(s)} \\\midrule
Bunny  &&&& 0.8337 && 17.4737 &&&& 0.8390 && 16.9883 &&&& 0.8217 && 14.1660 &&&& \bfseries 0.8202 && \bfseries 12.6767 \\
Dragon &&&& 101.0971 && 55.2926 &&&& 4.2504 && 31.3989 &&&& 46.2645 && 43.9151 &&&& \bfseries 0.5024 && \bfseries 15.2778\\
Buddha &&&& 0.7196  && 88.1237 &&&& 0.1838 && 53.9708 &&&& 0.1636 && 44.7148 &&&& \bfseries 0.1627  && \bfseries 43.3784\\
Chef &&&& 1.9897  && 246.6610 &&&& 0.2840 && 97.2748 &&&& 0.8841 && 161.4819 &&&& \bfseries 0.2699  && \bfseries 91.0270\\
Chicken &&&& 3.3135  && 74.9155 &&&& 0.4674 && 20.0214 &&&& 2.4075 && 61.5453&&&& \bfseries 0.4650  && \bfseries 19.4043\\
Trex &&&& 3.9160  && 130.9786 &&&& 0.3141 && 34.6767 &&&& 0.3018 && 138.4936 &&&& \bfseries 0.2942  && \bfseries 32.2257\\
Parasa &&&& 2.9139  && 70.2569 &&&& 0.4051 && 24.8153 &&&& 0.3916 && 63.6982 &&&& \bfseries 0.3717  && \bfseries 18.0444\\
\bottomrule
\end{tabular}}
\end{table*}

\begin{table*}[htbp]
\caption{Accuracy and efficiency comparison of all competed approaches, where small value indicates good performance and bold number denotes the best result.}
\label{tab:crosssection}
\ra{1.3}
\scalebox{.98}{
\begin{tabular}{@{}llcrrrcccrrrcccrrrcccrrr@{}}%
\toprule
&&& \multicolumn{3}{c}{MAICP~\cite{govindu2014averaging}}  &&&&\multicolumn{3}{c}{CFTrICP~\cite{Zhu2014Surface}}&&&&\multicolumn{3}{c}{LRS~\cite{arrigoni2016global}}&&&&\multicolumn{3}{c}{Ours}\\
\cmidrule{4-6} \cmidrule{10-12} \cmidrule{16-18} \cmidrule{22-24}
Datasets && Ini. \emph{Obj.} &  \emph{Obj.} && \emph{T(s)}  &&&&  \emph{Obj.} && \emph{T(s)} &&&&  \emph{Obj.} && \emph{T(s)} &&&&  \emph{Obj.} && \emph{T(s)}\\\midrule
Bunny  &&2.3103& \bfseries 0.8199 && 19.4722 &&&& 0.9516 && 23.4446 &&&& 0.8337 && 17.4738 &&&&  0.8217 && \bfseries 13.9604\\
Dragon && 4.3288 &  3.5527 &&81.9549 &&&& 2.5359 && 23.1322 &&&& 101.0971 && 55.2926 &&&& \bfseries 0.5024 && \bfseries 15.2029\\
Buddha && 3.1969 &  0.5617 &&  85.9106 &&&& 1.2196 && 56.5726 &&&& 0.7196  && 88.1237 &&&& \bfseries 0.1627 && \bfseries 44.9560\\
Chef && 2.0136 & 1.0235  && 120.4028 &&&& 0.6677 && 149.3099 &&&& 1.9897 && 246.6610 &&&& \bfseries 0.2699 && \bfseries 113.3068\\
Chicken && 1.4618 & \bfseries 0.4622  && 25.9324 &&&& 0.4934 && \bfseries 17.7066 &&&& 3.3135 && 74.9155 &&&& 0.4650 && 19.4043\\
Trex && 1.6461 & 0.7614  &&47.1791 &&&& 0.4966 && 62.1473 &&&& 3.9160 && 130.9786 &&&& \bfseries 0.2942 && \bfseries 32.2257\\
Parasa && 2.3179 & 0.7366  &&29.9331 &&&& 0.5846 && 51.7525 &&&& 2.9139 && 70.2569 &&&& \bfseries 0.3717 && \bfseries 18.0444\\
\bottomrule
\end{tabular}}
\end{table*}

Contrary to the intuition, the proposed approach is the most efficient among all variants of LRS decomposition based approaches.
Although some time is required by the matrix completing and weight calculation, it is only a small part of time spent on the multi-view registration. Usually, the most time-consuming operation is the establishment of point correspondence, which is included in pair-wise registration of LRS decomposition based approaches. For one special scan pair, fine initial parameters will cost less time to achieve accurate pair-wise registration. As shown in Fig. 1, pair-wise is the basis of multi-view registration, which provides the initial parameters for pair-wise registration in return. Since both matrix completing and motion weight can lead to robust and accurate multi-view registration, they can provide good initial parameters for pair-wise registration. Therefore, both matrix completing and motion weight can accelerate pair-wise registration, so they improve the efficiency of multi-view registration.

\subsection{Comparison}
To illustrate its performance, the proposed approach was compared with three state-of-the-art approaches, there are the motion averaging with the TrICP algorithm (MAICP)~\cite{govindu2014averaging}, the coarse to fine registration approach (CFTrICP) ~\cite{Zhu2014Surface}, and the original LRS decomposition based approach (LRS) ~\cite{arrigoni2016global}. Results of multi-view registration are also measured in the form of $Obj.$ and run time.

\begin{figure*}[htbp]
\centering
\centerline{\includegraphics[scale=0.44]{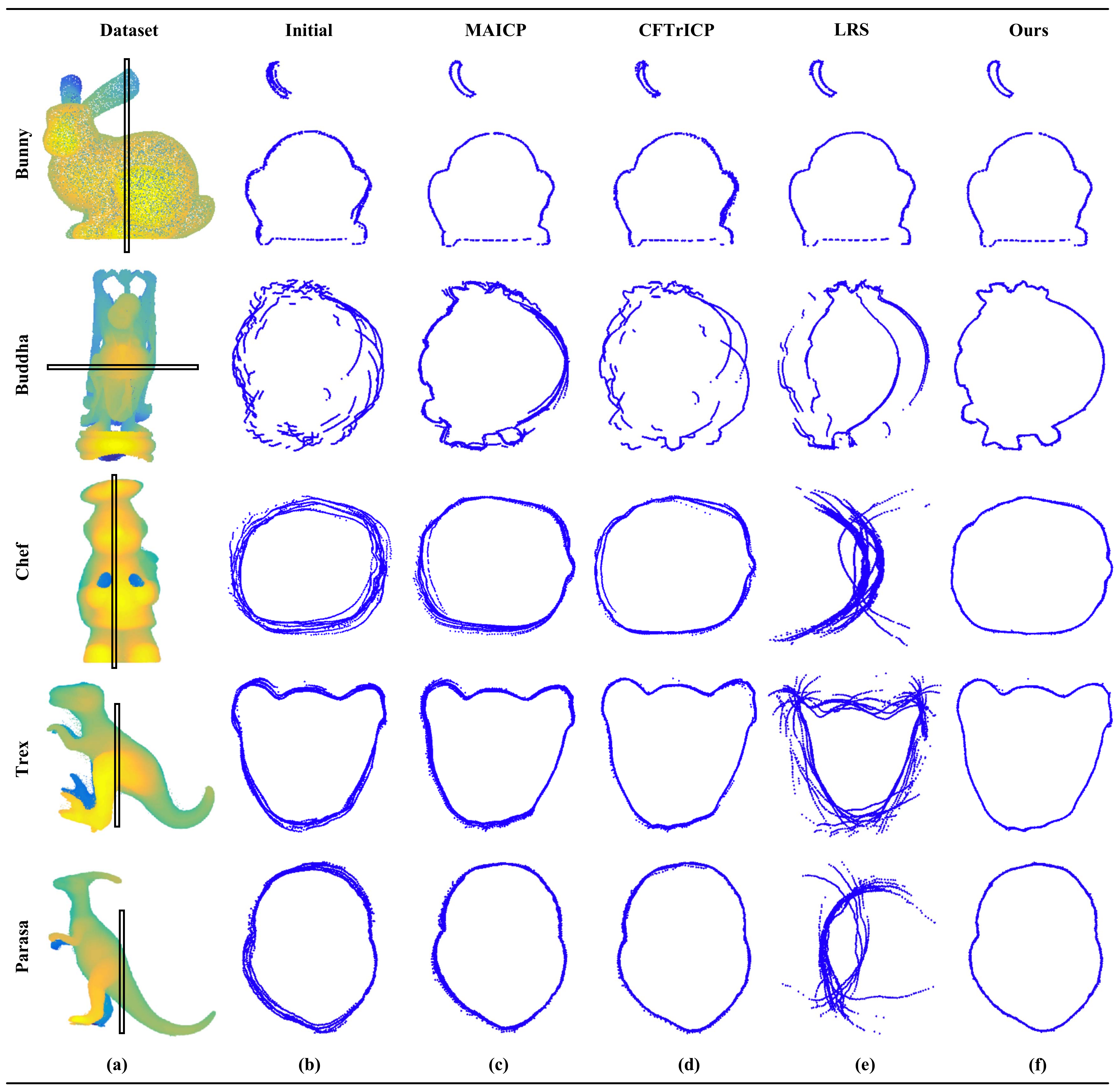}}
\caption{Comparison of different approaches in the form of cross-section. (a) Reconstructed 3D models. (b) Initialization. (c) MAICP. (d) CFTrICP. (e) LRS. (f) Ours. }\medskip
\label{fig:crosssection}
\end{figure*}

\subsubsection{Accuracy and efficiency}
For the comparison of accuracy and efficiency, experiments were carried on seven data sets with the same initial parameters. Comparison results of all competed approaches are displayed in Tab. ~\ref{tab:crosssection}. To illustrate the comparison in a more intuitive manner, Fig. ~\ref{fig:crosssection} displays the registration results of five data sets for all competed approaches in the form of cross-sections. As shown in Tab. ~\ref{tab:crosssection} and Fig. ~\ref{fig:crosssection}, the proposed approach can always obtain good results of multi-view registration. While, other approaches can not always achieve good multi-view registration.

As the pair-wise registration is the basis of multi-view registration, MAICP utilizes the TrICP algorithm to estimate relative motions of some scan pairs with high overlap percentages and then views these relative motions as the input of motion averaging algorithm to compute global relative motions. In MAICP, one unreliable relative motions will lead to inaccurate results, even other relative motions are very reliable. Therefore, only fine initial parameters can lead to good multi-view registration. With other global motions fixed, CFTrICP alternately refines each global motion by the TrICP algorithm, so its final registration result is always better than initial results. But this approach is easy to trap into local minimum. For good registration, CFTrICP requires to be provided with fine initial parameters. Otherwise, it is difficulty to achieve good registration.

As MAICP, LRS decomposition based approach also utilizes a set of relative motions to recover global motions for the multi-view registration. It is robust to unreliable relative motions but sensitive to the sparsity of reconstructed matrix. Without the matrix completion, the reconstructed matrix are always sparse, which will lead to the failure of LRS decomposition. By introducing the matrix completion, the proposed approach
can always obtain the robust LRS decomposition results for multi-view registration. What's more, the weight of relative motions arrows the LRS decomposition to pay more attention to reliable relative motions, which can further improve the performance of LRS decomposition for the multi-view registration. Therefore, the proposed approach can almost obtain the best registration results among all competed approaches.

\subsubsection{Robustness}

To compare the robustness, all competed approaches were tested on Stanford
Dragon with different groups of initial parameters, which were acquired by adding some uniformly random noises to the rotation matrix. To eliminate the randomness, 20 Monte Carlo (MC) trials were carried out with respect to each noise level. For comparison,
mean value of $Obj.$ and run time are displayed in Tab.~\ref{tab: GrtCom}. As shown in Tab.~\ref{tab: GrtCom}, the proposed approach obtain the most accurate registration results under varied noise levels. Although all other approaches can obtain good registration results under low noise level, their performance will decrease seriously with the increase of noise level.

\begin{table*}[htb!]
\caption{Robustness comparison of all competed approaches, where small value indicates good performance and bold number denotes the best result.}
\label{tab: GrtCom}
\ra{1.3}
\scalebox{0.975}{
\begin{tabular}{@{}llrrrrrcrrrrrcrrrrrcrrrrr@{}}
\toprule
&&& \multicolumn{5}{c}{MAICP~\cite{govindu2014averaging}} && \multicolumn{5}{c}{CFTrICP~\cite{Zhu2014Surface}} &&\multicolumn{5}{c}{LRS~\cite{arrigoni2016global}}  &\multicolumn{5}{c}{Ours}\\
\cmidrule{5-7} \cmidrule{11-13} \cmidrule{17-19}  \cmidrule{23-25}
Noise level &&&& \emph{Obj.} && \emph{T(s)} &&&& \emph{Obj.}  && \emph{T(s)}   &&&& \emph{Obj.} && \emph{T(s)}   &&&& \emph{Obj.} && \emph{T(s)} \\\midrule
$\left[ { - 0.02 ,0.02 } \right] (rad)$   &&&&0.5481 && 14.9105 &&&& 0.5227 && 18.2295 &&&& 0.5639  && 37.4798  &&&& \bfseries 0.5052 &&\bfseries 14.1806\\
$\left[ { - 0.04 ,0.04 } \right] (rad)$   &&&& 0.5542 && 21.1851  &&&&0.6460 &&42.5968 &&&& 3.5932 && 124.2523 &&&& \bfseries 0.5088 && \bfseries 17.8799 \\
$\left[ { - 0.06 ,0.06 } \right] (rad)$  &&&& 3.4704 && 132.4142 &&&& 1.1665 && 40.7482 &&&& 3.3980 &&146.8945 &&&& \bfseries 0.5101 && \bfseries 33.0372 \\
$\left[ { - 0.08 ,0.08 } \right](rad)$  &&&& 3.6264 && 171.5397 &&&& 1.3439 && \bfseries 44.1234  &&&& 3.8113 && 151.7663 &&&& \bfseries 0.5114 &&  58.2175\\
$\left[ { - 0.10 ,0.10 } \right](rad)$  &&&& 3.6270 && 166.5593 &&&& 1.8182 && \bfseries 49.3438 &&&& 52725 && 161.9022 &&&& \bfseries 0.6627 && 60.6974\\
\bottomrule
\end{tabular}}
\end{table*}

Similar to LRS decomposition based approach, MAICP also recovers all global motions form a set of relative motions, which estimated by the TrICP algorithm. However, this approach is sensitive to unreliable relative motions and one unreliable relative motion will lead to the failure of multi-view registration. Under high noise level, it is difficulty to accurately estimate the overlap percentage of each scan pair, which will certainly introduce the unreliable pair-wise registration. Hence, the performance of MAICP turn to be seriously decreased. Different from MAICP, CFTrICP utilizes the TrICP algorithm to refine each global motion alternately, which make it easy to trap into local minimum. Under low noise level, initial global motions are accurate and they can be easily refined. However, with the increase of noise level, CFTrICP may be convergent to local minimum quickly due to inaccurate initial parameters and global motions are diffculty to be refinded.

Although LRS decomposition based approach is robust to a small portion of unreliable relative motions, it is sensitive to the sparsity of the reconstructed matrix. Under low noise level, a set of reliable relative motions are available to reconstruct the matrix for LRS matrix decomposition, which may result in good multi-view registration. With the increase of noise level, some available relative motions turn to be unreliable, which can reduce the sparsity of the reconstructed matrix and lead to the failure of multi-view registration. By introducing the matrix completion, the sparsity of the reconstructed matrix is reduced, which increase the robustness of LRS decomposition. Besides, the weight of relative motions allows the LRS decomposition pay more attention to these reliable relative motions. Therefore, the proposed approach can achieve multi-view registration with good performance even under high noise levels.

\section{Conclusions}\label{conclusions}

This paper proposes a novel approach for multi-view registration based on the weighted LRS matrix decomposition. According to the anti-symmetry property of relative motions, it firstly applies the completion strategy to reduce the sparsity of reconstructed matrix to be decomposed. As the LRS decomposition algorithm is sensitive to the sparsity of potentially decomposed matrix, the completion strategy can improve its robustness. Additionally, it introduces the weight to indicate the reliability of each block element of the reconstructed matrix and then proposes the weighted LRS decomposition algorithm. This algorithm can pay more attention to reliable block elements with large weight and achieve more accurate multi-view registration. Besides, compared with the original LRS decomposition, the proposed approach can also make the progress in efficiency for multi-view registration. Experiments on public available data sets demonstrate its good performance over the state-of-the-art approaches on robustness, accuracy, and efficiency.

\begin{figure}[htbp]
\centering
\begin{minipage}[b]{0.4\linewidth}
  \centering
  \centerline{\includegraphics[scale=0.45]{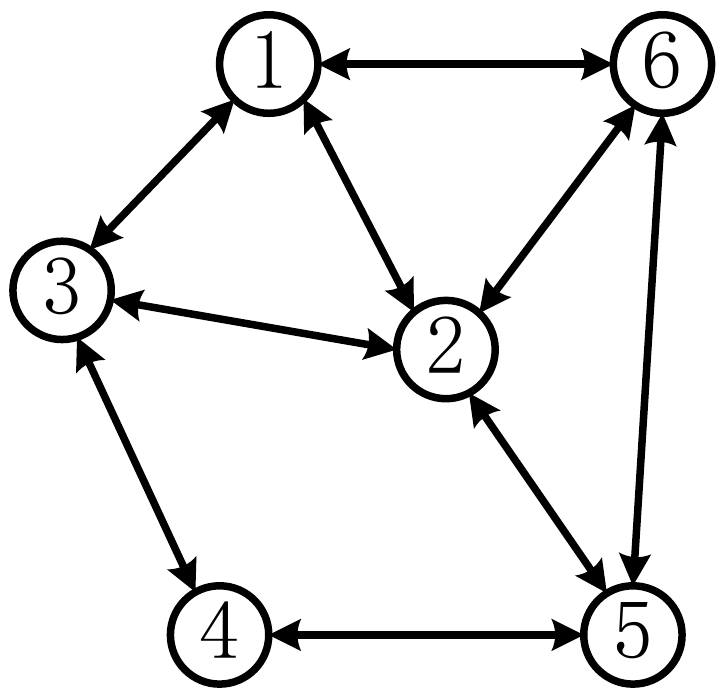}}
  \centerline{(a)}
\end{minipage}
\qquad \qquad
\begin{minipage}[b]{0.4\linewidth}
  \centering
  \centerline{\includegraphics[scale=0.45]{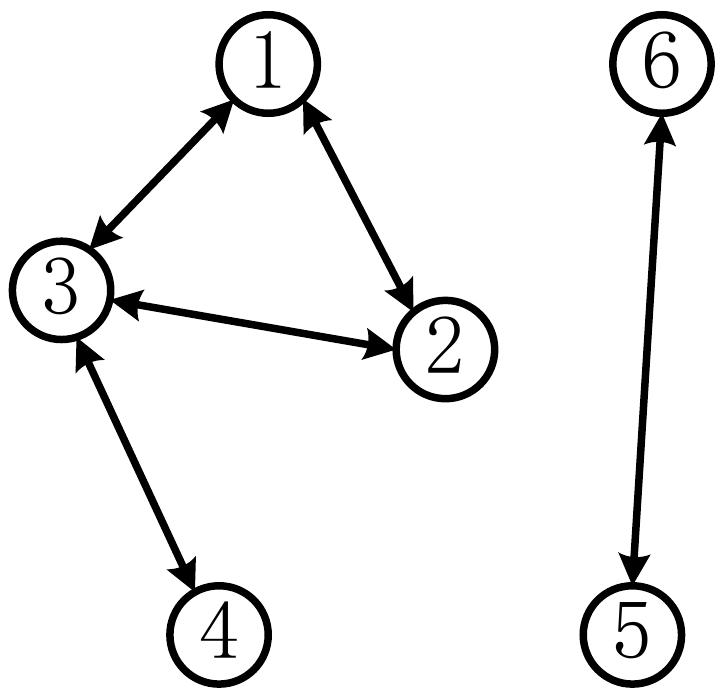}}
  \centerline{(b)}
\end{minipage}
\caption{ Multi-view range scans are transformed into a model graph, where each circle indicates one range scan and each line with arrow denotes two available relative motions (one of them may be obtained by the anti-symmetry property). (a) Complete model. (b) Partial models.}\medskip
\label{fig: Limi}
\end{figure}

Although the proposed approach has good performance for the multi-view registration, it does not mean that this approach can solve any multi-view registration problem. As shown in Fig. \ref{fig: Limi}, multi-view range scans are transformed into a model graph, where each circle indicates one range scan and each line with arrow denotes one available relative motion. Actually, the proposed approach can only achieve the multi-view registration of these range scans, which can be transformed into a completed model. It is not suitable for the multi-view registration of these range scans, which can only be denoted by several partial models. However, it should be noted that many approaches for multi-view registration proposed so far share this limitation as well.

Similar to most of multi-view registration approaches, the proposed approach should be provided with initial global motions. Therefore, our future work will focus on the estimation of initial global motions for the multi-view registration.

\section*{Acknowledgements}

This work is supported by the National Natural Science Foundation of China under Grant No.
61573273 and Natural Science Foundation of Jiangsu Province under Grant No. BK20161516. It is also supported by State Key Laboratory of Rail Transit Engineering Informatization (FSDI) under Grant No. SKLK16-09. Besides, we would like to thank Federica Arrigoni for providing the MATLAB implementation of~\cite{arrigoni2016global}.


\bibliographystyle{IET_journal}
\bibliography{sample}

\end{document}